%% file: egpaper_for_review.tex
\documentclass[10pt,twocolumn,letterpaper]{article}

\usepackage{iccv}
\usepackage{times}
\usepackage{epsfig}
\usepackage{graphicx}
\usepackage{amsmath}
\usepackage{amssymb}

\usepackage{datetime2}
\usepackage{graphicx}
\usepackage{amsmath}
\usepackage{amssymb}
\usepackage{booktabs}
\usepackage{mathtools}
\usepackage{amsthm}
\usepackage{eucal}
\usepackage{dsfont}
\usepackage{bm} %added 
\usepackage{pifont}% 
\newcommand{\cmark}{\ding{51}}%
\newcommand{\xmark}{\ding{55}}%

\DeclareMathOperator*{\argmin}{arg\,min}
\DeclareMathOperator*{\argmax}{arg\,max} 

\DeclareMathOperator*{\argminC}{\arg\min}

\usepackage{color, colortbl}
\definecolor{Gray}{gray}{0.9}

\newcommand{\ceq}{\stackrel{\mathclap{\normalfont\mbox{c}}}{=}}
\DeclareMathAlphabet{\mathcal}{OMS}{cmsy}{m}{n}

\def \all {{\mathtt{ALL}}}

\def \calD{{\mathcal{D}}}
\def \calZ{{\mathcal{Z}}}
\def \calY{{\mathcal{Y}}}
\def \calH{{\mathcal{H}}}
\usepackage[pagebackref=true,breaklinks=true,letterpaper=true,colorlinks,bookmarks=false]{hyperref}

\usepackage[capitalize]{cleveref}
\crefname{section}{Sec.}{Secs.}
\Crefname{section}{Section}{Sections}
\Crefname{table}{Table}{Tables}
\crefname{table}{Tab.}{Tabs.}
\iccvfinalcopy % *** Uncomment this line for the final submission

\def\iccvPaperID{11198} % *** Enter the ICCV Paper ID here

\ificcvfinal\fi

\begin{document}

%%%%%%%%% TITLE
\title{Parametric Information Maximization for Generalized Category Discovery}

\author{Florent Chiaroni \thanks{Corresponding author: florent.chiaroni.1@etsmtl.net} \\
\'ETS Montreal \& Thales cortAIx\\
Montreal, Canada\\
%{\tt\small florent.chiaroni.1@etsmtl.net}
\and
Jose Dolz\\
\'ETS Montreal\\
Montreal, Canada\\
\and
Ziko Imtiaz Masud\\
Thales CortAIx\\
Montreal, Canada\\
\and
Amar Mitiche\\
INRS\\
Montreal, Canada\\
\and
Ismail Ben Ayed\\
\'ETS Montreal\\
Montreal, Canada\\
}

\maketitle
\ificcvfinal\thispagestyle{empty}\fi

%%%%%%%%% ABSTRACT
\begin{abstract}
We introduce a Parametric Information Maximization (PIM) model for the Generalized Category Discovery (GCD) problem. Specifically, we propose a bi-level optimization formulation, which explores a parameterized family 
of objective functions, each evaluating a weighted mutual information between the features and the latent labels, subject to supervision constraints from the labeled samples. Our formulation mitigates the class-balance bias encoded in standard information maximization approaches, thereby handling effectively both short-tailed and long-tailed data sets. We report extensive experiments and comparisons demonstrating that our PIM model consistently sets new state-of-the-art performances in GCD across six different datasets, more so when dealing with challenging fine-grained problems. 
%Our code: \url{}
\end{abstract}

%%%%%%%%% BODY TEXT
\section{Introduction}
\label{sec:intro}

Deep learning methods are driving progress in a wide span of computer vision tasks, particularly when large labeled datasets are easily accessible for training. Obtaining such large datasets is a cumbersome process, which is often a limiting factor impeding the scalability of these models. To alleviate this limitation, semi-supervised learning (SSL) has emerged as an appealing alternative, which leverages both labeled and unlabeled data to boost the performance of deep models. Despite recent success, SSL approaches work under the \textit{closed-set} assumption, in which the categories in the labeled and unlabeled subsets share the same underlying class label space. Nevertheless, this assumption rarely holds in real scenarios, where novel categories may emerge in conjunction with known classes, which typically results in significant drops in the performances of standard supervised deep learning models. Thus, the ability to detect whether the input of a deep learning model belongs or not to a set of \textit{known} classes seen during training is essential for robust deployment in a breadth of critical application areas, such as, medicine, security, finance, agriculture, marketing, and engineering \cite{bendale2016towards, Macedo2021}. %roy2022
Thus, devising novel learning models that can address the realistic \textit{open-set} scenario is of paramount importance.

Novel category discovery (NCD) \cite{han2019learning,han2020automatically} tackles this problem by exploiting the knowledge learned from a set of relevant known classes to improve the clustering of the unknown categories. Nevertheless, NCD assumes the two sets of classes to be disjoint, which means that the unlabeled dataset contains only instances belonging to the set of novel categories. {\em Generalized category discovery (GCD)} \cite{vaze2022generalized} considers a more general scenario, where unlabeled data contain instances from both seen and novel classes. This scenario is particularly challenging, as learning is performed under class distribution mismatch, and the unlabeled data may contain categories never encountered in the available labeled set.

\paragraph{Contributions:} In this work, we address the generalized category discovery task from an information-theoretic perspective. Our contributions are
summarized as follows:

\begin{itemize}
    \item We introduce a \textit{Parametric Information Maximization} (PIM) model for GCD.
    Specifically, we propose a bi-level optimization formulation, which explores a parameterized family of objective functions, each evaluating a weighted mutual information between the features and the latent labels, subject 
    to supervision constraints from the labeled samples. Our formulation mitigates the class-balance bias encoded in standard information maximization, deals effectively with both short-tailed and long-tailed data sets, and 
    is model-agnostic (i.e., could be used in conjunction with any feature extractor). 

    \item We report extensive experiments and comparisons demonstrating that PIM consistently sets new state-of-the-art performances across six different datasets, with larger gaps on the more challenging fine-grained benchmarks. It outperforms both specialized GCD methods and standard information-maximization approaches.
    
\end{itemize}

\section{Related work}
\label{sec_related_work}

\textbf{Semi-supervised learning (SSL)} has been widely explored in the machine learning and computer vision community. This learning paradigm aims at leveraging large unlabeled datasets that contain the same set of classes as the labeled samples. Due to their satisfactory performance, consistency-based approaches have gained popularity recently, such as Mean-Teacher \cite{tarvainen2017mean}, MixMatch \cite{berthelot2019mixmatch}, UDA \cite{xie2020unsupervised} or FixMatch \cite{sohn2020fixmatch}. An interesting alternative is self-training, which relies on the generation of pseudo-labels from a small amount of labeled data \cite{rizve2020defense,zheng2022simmatch}, or in solving surrogate classification tasks \cite{gidaris2018unsupervised,zhai2019s4l}. Nevertheless, the main limitation is that most existing SSL models rely on the {\em closed-set} assumption, as they do not consider unlabeled data points sampled from novel semantic categories.

\textbf{Novel Class Discovery (NCD)}, which was formalized in \cite{han2019learning}, relaxes the closed-set assumption, as it focuses on discovering new categories in the unlabeled set by leveraging the knowledge learned from the labeled set. AutoNovel \cite{han2021autonovel} (also referred to as RankStats) resorts to ranking statistics as an efficient approach for NCD. First, a good embedding is learned in a self-supervised manner for learning the early feature representation layers, which is followed by a supervised fine-tuning step with labeled samples for learning high level feature representations. Finally, to determine whether two instances from the unlabeled set are from the same category, a robust ranking statistics approach is introduced. A dual ranking statistics method coupled with mutual knowledge distillation is further proposed in \cite{zhao2021novel}. OpenMix \cite{zhong2021openmix} showed that mixing up both labeled and unlabeled data can prevent the representation learning model from overfitting the labeled categories. Some other methods \cite{jia2021joint,zhong2021neighborhood} adopt contrastive learning for the novel category discovery task. UNO \cite{fini2021unified} unifies a cross-entropy loss to jointly train the model with both the labeled and unlabeled data. Despite the good performance in discovering new categories, these methods assume that the test dataset only contains instances from the novel classes. A recent work by \cite{zhang2022mutual} presented a method based on a mutual-information measure, which is different from the discriminative and constrained mutual information we introduce in this work.
The mutual information in \cite{zhang2022mutual} evaluates the relation between the old and novel categories in the label space, arguing that maximizing such a measure promotes transferring semantic knowledge. In our case, we introduce a 
parametric, bi-level optimization of the mutual information between the feature and label spaces, on both labeled and unlabeled samples.

\textbf{Generalized Category Discovery (GCD)} extends NCD by allowing both old and new classes to coexist in the unlabeled dataset, which we tackle in this work. 
This pragmatic yet challenging scenario was recently introduced in \cite{vaze2022generalized} and triggered several other recent studies of the GCD problem. In \cite{vaze2022generalized}, the authors proposed to fine-tune a pre-trained DINO ViT \cite{caron2021emerging} with one supervised and one self-supervised contrastive term. Then, they used a semi-supervised clustering for label assignment. Note that, while UNO \cite{fini2021unified} and RankStats \cite{han2021autonovel} are originally investigated for the NCD task, they are adapted for GCD in the recent study in \cite{vaze2022generalized}, yielding UNO+ and RankStats+, respectively. Another recent approach, referred to as ORCA \cite{cao2022openworld}, addressed a similar problem, naming it \textit{open world semi-supervised learning}. ORCA consists of controlling the intra-class variance of the seen classes to align and reduce the learning gap w.r.t. novel categories.

\textbf{Maximizing the mutual information}. Our discriminative partitioning approach (PIM) is built on the general and well-known {\em InfoMax} principle \cite{linsker1988self}, which prescribes maximizing the mutual information (MI) between the inputs and outputs of a system. Several variants of this general principle have been recently used in machine learning and computer vision tasks, including deep clustering \cite{jabi2019deep,krause2010discriminative,hu2017learning}, few-shot learning \cite{boudiaf2020information}, representation learning \cite{tschannen2019mutual,hjelm2018learning,bachman2019learning,kemertas2020rankmi}, deep metric learning \cite{boudiaf2020unifying} and domain adaptation \cite{pan2020exploring}. To the best of our knowledge, addressing the GCD problem from an information-theoretic perspective remains unexplored.

The pioneering discriminative clustering model in \cite{krause2010discriminative} and the recent transductive few-shot method in \cite{boudiaf2020information} 
are closely related to our work, as they both maximize the mutual information between the features and the latent labels. However, as we shall see in our 
experiments, the direct application of information maximization \cite{krause2010discriminative,boudiaf2020information} to GCD may not be highly competitive. First, the standard mutual-information objective has a strong encoded bias 
towards balanced partitions, via its marginal-entropy term, which might be detrimental to performances. In this work, we introduce a parametric family of mutual-information objectives, which we tackle with a bi-level optimization formulation, thereby estimating automatically the weight of the marginal-entropy term. Our parametric information maximization effectively deals with both short-tailed and long-tailed data sets, mitigating the class-balance bias. Secondly, the InfoMax models 
in \cite{krause2010discriminative, boudiaf2020information} were designed in the scenario where the unlabeled set contains examples from the classes seen in the available labeled set. Finally, in \cite{krause2010discriminative, boudiaf2020information}, the mutual-information objective is defined over the set of unlabeled samples. In contrast, we propose a constrained mutual-information formulation defined over both labeled and unlabeled samples, thereby capturing the distribution of the whole data set in the context of GCD. As we will see in our experiments, our parametric, bi-level information maximization substantially outperforms \cite{krause2010discriminative,boudiaf2020information} in the GCD scenario.  

\section{Generalized Category Discovery problem}
\label{subsec_GCD_problem_formulation}

\noindent \textbf{Problem definition.} Assume we are given a dataset $\calD$ composed of two subsets so that $\calD= \calD_L \cup \calD_U$. First, $\calD_L=\{(\bm{x}_i,\bm{y}_i)\}_{i=1}^{N}$ refers to a labeled subset containing $N$ images from a set of known classes in $\calY_L$. For each image $\bm{x}_i$ in $\calD_L$, we have access to its corresponding one-hot vector label $\bm{y}_i= (y_{i,k})_{1 \leq k\leq K^{\text{old}}}$, where $K^{\text{old}}=|\calY_L|$ is the number of classes in $\calY_L$. $y_{i,k} = 1$ if $\bm{x}_i$ belongs to class $k$, and $0$ otherwise. Now, let $\calD_U=\{\bm{x}_i\}_{i=1}^{M}$ denote the unlabeled subset, which contains $M$ images from a set of classes $\calY_U$ composed of {\em known} classes, as well as {\em novel} classes, i.e., $\calY_L \subset \calY_U$. Note that, during inference, $K=|\calY_U|$ is the total number of classes, which contains both known and novel categories. Given this setting, the Generalized Category Discovery (GCD) task introduced in \cite{vaze2022generalized} consists in partitioning
the images in the unlabeled set into separate clusters at test time. Each obtained cluster is supposed to represent a separate known or novel category. In other words, the GCD problem amounts to jointly solving (i) a semi-supervised classification task for the known classes; and (ii) a clustering task for the novel classes.

\noindent \textbf{Notation.} Let us denote $g_{\bm{\theta}}: \calD \rightarrow \calZ \subset \mathbb{R}^D$ as the trained encoder responsible for mapping an input image $\bm{x}_i$ into a feature vector $\bm{z}_i$ of dimension $D$, with $\theta$ the set of trainable parameters and $\calZ$ the set of all embedded features, for both the labeled and unlabeled samples. We now define a soft partitioning model $f_{\bm{W}}: \calZ \rightarrow [0,1]^K$, which is parameterized by weight matrix $\bm{W}=(\bm{w}_k)_{1 \leq k \leq K}$, where $\bm{w}_k=(w_{k,n})_{1 \leq n \leq D}$ denote its trainable parameters. For each input feature map $\bm{z}_i$, $f_{\bm{W}}$ outputs a softmax prediction vector $\bm{p}_i=(p_{i,k})_{1 \leq k \leq K}$ of dimension $K$, defined on the standard ($K-1$)-probability simplex domain $\Delta^{K-1}= \{\bm{p}_i \in [0,1]^K~|~ \bm{p}_i^T \bm{1} = 1\}$. 
Note that, similarly to the prior work in \cite{vaze2022generalized}, we assume the number of clusters during the partitioning task to be known.

Let $Z \in \mathbb{R}^{D}$ denote a random variable representing the feature map. $Z$ follows $\mathbb{P}(Z)$, which denotes the distribution of the set of embedded features $\calZ$. Hence, each feature map data point $\bm{z}_i$ is a realization of $Z$. Furthermore, let $Y\in \calY=\{1,\dots,K\}$ be the random variable following the dataset label distribution $\mathbb{P}(Y)$.

\section{Background on information maximization}

\paragraph{Marginal distributions.} Let $\bm{\pi}=(\pi_k)_{1 \leq k \leq K}$, where $\pi_k=\mathbb{P}(Y=k;\bm{W})$ denote the marginal distributions that one can approximate by the soft\footnote{We use the term \textit{soft} because the proportions are directly estimated on the softmax predictions, instead of using hard labels.} proportion of points within each cluster, via Monte-Carlo estimation, as follows:

\begin{equation}
    \begin{aligned}
        \pi_k &= \int_{\bm{z}} \mathbb{P}(Z=\bm{z}) \mathbb{P}(Y=k|Z=\bm{z};\bm{W})\text{d}\bm{z} \\
        & \approx \frac{1}{|\mathcal{Z}|} \sum_{i \in \calZ} \mathbb{P}(Y=k|Z=\bm{z}_i;\bm{W})= \frac{1}{|\mathcal{Z}|} \sum_{i \in \calZ} p_{i,k}
    \end{aligned}
    \label{eq_marg_prob}
\end{equation}

\paragraph{Mutual Information.} The mutual information between the labels and the features maps can be written as follows:
\begin{equation}
\label{eq_mutual_info}
    I(Y,Z) = \mathcal{H}(Y) - \mathcal{H}(Y|Z),
\end{equation}
with $\mathcal{H}(Y)$ referring to the entropy of the marginal distributions $\mathbb{P}(Y=k;\bm{W})$, and $\mathcal{H}(Y|Z)$ referring to the entropy of the conditional probability distribution $\mathbb{P}(Y|Z; \bm{W})$.

\paragraph{Marginal entropy.} The marginal entropy term  $\mathcal{H}(Y)$ in \eqref{eq_mutual_info} can be estimated, w.r.t. the soft marginal distributions approximation in \eqref{eq_marg_prob}, as follows:

\begin{equation}
\label{eq_mutual_ent}
    \begin{aligned}
        \mathcal{H}(Y) &= -\sum_{k=1}^K \mathbb{P}(Y=k;\bm{W}) \log \mathbb{P}(Y=k;\bm{W}) \\
        &= -\sum_{k=1}^K \pi_k \log \pi_k
    \end{aligned}
\end{equation}

\noindent \textbf{Weakness of InfoMax based approaches \cite{boudiaf2020information,krause2010discriminative}}.
It is common in the literature to maximize the unsupervised mutual information in Eq. \eqref{eq_mutual_info}, which is often defined over unlabeled samples. This is the case, for instance, of the transductive few-shot inference in \cite{boudiaf2020information} (TIM) and the discriminative clustering in \cite{krause2010discriminative} (RIM). 
A closer look at the marginal-entropy term in \eqref{eq_mutual_ent} enables to write it, up to a constant, as a Kullback-Leibler (KL) divergence between the marginal probabilities of predictions and the uniform distribution:

        \begin{align}
            \mathcal{H}(Y) & \ceq %\log (K)
            - \mathcal{D}_{KL}(Y||\mathcal{U}_K), \label{eq:link_entropy_kl}
        \end{align}
where $\ceq$ stands for equality up to additive and/or non-negative multiplicative constant, and $\mathcal{U}_K$ is the uniform distribution over K classes. Thus, the term $\mathcal{H}(Y)$ pushes the marginal distribution towards the uniform distribution, as made explicit by the previous equation, thereby encoding a strong bias towards balanced partitions. Note that this standard mutual information objective lacks a mechanism to explicitly control the weight of the marginal entropy. Therefore, this term has the potential to harm the performance in the case of imbalanced scenarios, where the underlying class distribution is no longer uniform. Based on the above-identified limitation of the mutual information, we introduce a parametric family of mutual-information objectives, which we tackle with a bi-level optimization formulation, thereby estimating the relative weight of the marginal-entropy term.

\section{Proposed bi-level and constrained InfoMax} \label{subsec_bilevel_optim}

\paragraph{Constrained mutual information} We propose to maximize a constrained version of the mutual information presented in \eqref{eq_mutual_info}, integrating supervision constraints on the conditional probabilities $\bm{p}_i$ of the samples within the labeled set. Our constrained information maximization reads:
\begin{equation}
        \max_{\bm{W}} \, \mathcal{H}(Y) - \mathcal{H}(Y | Z) \quad \mbox{s.t.} \quad \bm{y}_i=\bm{p}_i \quad \forall \bm{z}_i \in \calZ_L
    \label{eq_constrained_mutual_info}
\end{equation}
where $\calZ_L$ denotes the set of embedded features for the labeled samples.   
It is straightforward to notice that by plugging the equality constraints in \eqref{eq_constrained_mutual_info} into the mutual information, one could write the objective as follows:
\begin{equation}
    \begin{aligned}
        \min_{\bm{W}} \, \sum_{k=1}^K \pi_k \log \pi_k 
    - \frac{1}{|\calZ|} \sum_{i \in \calZ} \sum_{k=1}^K h_{i,k} \log p_{i,k},
    \end{aligned}
    \label{eq_auxiliary}
\end{equation}
where $h_{i,k}=y_{i,k}$ if $\bm{z}_i \in \mathcal{Z}_L$ and $h_{i,k} = p_{i,k}$ otherwise. That is, for  $\bm{y}_i=\bm{p}_i \quad \forall \bm{z}_i \in \calZ_L$, the objectives in
\eqref{eq_constrained_mutual_info} and \eqref{eq_auxiliary} are equal to each other. 
Interestingly, the terms corresponding to $h_{i,k}=y_{i,k}$ in \eqref{eq_auxiliary} yield the standard 
cross-entropy (CE) loss for the labeled samples. This CE loss could be viewed as a {\em penalty} function for imposing 
constraints $\bm{y}_i=\bm{p}_i \quad \forall \bm{z}_i \in \calZ_L$, as it reaches its minimum when these constraints are satisfied. Therefore, we do not need to impose explicitly the equality constraints in \eqref{eq_constrained_mutual_info}.
Notice that, for the labeled samples, CE in \eqref{eq_auxiliary} replaced the conditional entropy term in the mutual information. This is reasonable as CE enables to jointly impose the supervision constraints while encouraging implicitly confident predictions, as it pushes them toward one vertex of the simplex. Both CE and conditional entropy reach their minima at the vertices of the simplex.  

\paragraph{Bi-level optimization} To mitigate the bias of the mutual information towards balanced partitions, we propose to explore a family of weighted versions of the objective in \eqref{eq_auxiliary}, which we parameterize with a variable parameter $\lambda$ and tackle as a bi-level optimization problem: 
\begin{equation}
    \begin{aligned}
F(\bm{W}, \lambda) = \underbrace{\sum_{k=1}^K \pi_k \log \pi_k}_{ \calH(Y)} \underbrace{-\frac{1}{|\calZ_L|} \sum_{i \in \calZ_L} \sum_{k=1}^K y_{i,k} \log p_{i,k}}_{\mbox{{\tiny CE}}} \\
      - \underbrace{\frac{\lambda}{|\calZ_U|} \sum_{i \in \calZ_U} \sum_{k=1}^K p_{i,k} \log p_{i,k}}_{\propto \calH(Y | Z)}
    \end{aligned}
    \label{eq_classifier_loss}
\end{equation}
where $\calZ_U$ denotes the set of embedded features for the unlabeled samples (i.e., $\calZ = \calZ_U \cup \calZ_L$).  
 Variable $\lambda \in (0,1]$ controls the effect of the unsupervised loss terms in Eq. \eqref{eq_classifier_loss}, i.e., confidence vs. class balance. Therefore, as we will see in 
 our experiments, learning $\lambda$ from the labeled set, via a bi-level optimization, yields highly competitive performances in the GCD setting, more so when dealing with long-tailed 
 (imbalanced) data sets. Our bi-level formulation reads:
\begin{equation}
\label{bi-level-formulation}
\min_{\bm{W}} F(\bm{W}, \lambda) \quad \mbox{s.t} \quad \lambda \in \argmax_{\lambda \in (0,1]} A_L (\lambda), 
\end{equation}
where $F$ is the upper-level objective and $A_L$ is the lower-level objective defined by the clustering accuracy\footnote{We used the Hungarian algorithm to align labels of the most consistent $K^{\text{old}}$ clusters (among the total $K$ clusters) with the $K^{\text{old}}$ class labels.} on the set of labeled samples: 
\begin{equation}
    A_L(\lambda) = \frac{1}{|\calZ_L|} \sum_{i=1}^{|\calZ_L|} \mathds{1}_{\{\hat{\bm{y}}_i (\lambda) = \bm{y}_i\}},
\end{equation}
and $\hat{\bm{y}}_i (\lambda)$ are the one-hot vector predictions on labeled samples maximizing parametric mutual information: 
\begin{equation}
\label{mutual-info-lambda}
G(\bm{W}, \lambda) = \sum_{k=1}^K \pi_k \log \pi_k - \frac{\lambda}{|\calZ|} \sum_{i \in \calZ} \sum_{k=1}^K p_{i,k} \log p_{i,k}
\end{equation}
To tackle our problem, we explore a finite set\footnote{In our experiments, we used $19$ values of $\lambda$ in $[5e^{-2}, 1]$, i.e. the cardinalty of set $\bm{\lambda}$ is $19$.} $\bm{\lambda}$ of uniformly-spaced values 
of variable $\lambda$ in $(0, 1]$. For each of these values of $\lambda$, we optimize $G(\bm{W}, \lambda)$ in Eq. \eqref{mutual-info-lambda} w.r.t to linear-classifier parameters $\bm{W}$ via standard gradient steps, thereby obtaining predictions 
$\hat{\bm{y}}_i (\lambda)$. Note that, although we explore several values of $\lambda$, this remains computationally efficient as the feature encoder parameters are fixed and only classifier parameters $\bm{W}$ are 
updated. For initializing $\bm{W}$, which could be viewed as class prototypes, we use the K-means++ algorithm. This process yields a prediction of the optimal $\lambda$ 
as: $\lambda_{\mbox{opt}} = \argmax_{\lambda \in \bm{\lambda}} A_L (\lambda)$. Finally, the partitioning solution of our GCD problem in Eq. \eqref{bi-level-formulation} is obtained by optimizing the upper-level objective
$F(\bm{W}, \lambda_{\mbox{opt}})$ via gradient steps.

\section{Experiments}

\begin{table*}[ht]
 \addtolength{\tabcolsep}{-3pt}
                \vskip -0.15in
                \begin{center}
                \begin{small}
                \begin{sc}
                \resizebox{0.7\textwidth}{!}{%
                \begin{tabular}{lcccccccccc}
                    \toprule
                     & & \multicolumn{3}{c}{CUB} & \multicolumn{3}{c}{Stanford Cars} & \multicolumn{3}{c}{Herbarium19} \\
                    \cmidrule(r){3-5} \cmidrule(r){6-8} \cmidrule(r){9-11} Approach & InfoMax & All & Old & New & All & Old & New & All & Old & New \\
                    \midrule
                    k-means &  & 34.3 & 38.9 & 32.1 & 12.8 & 10.6 & 13.8 & 12.9 & 12.9 & 12.8 \\
                    RankStats+ \cite{han2021autonovel} (TPAMI-21) &  & 33.3 & 51.6 & 24.2 & 28.3 & 61.8 & 12.1 & 27.9 & 55.8 & 12.8 \\
                    UNO+ \cite{fini2021unified} (ICCV-21) &  & 35.1 & 49.0 & 28.1 & 35.5 & \textbf{70.5} & 18.6 & 28.3 &  53.7 & 14.7 \\
                    ORCA \cite{cao2022openworld} (ICLR-22) &  & 27.5 & 20.1 & 31.1 & 15.9 & 17.1 & 15.3 & 22.9 & 25.9 & 21.3 \\ % ResNet50
                    ORCA \cite{cao2022openworld} - ViTB16 &  & 38.0 & 45.6 & 31.8 & 33.8 & 52.5 & 25.1 & 25.0 & 30.6 & 19.8 \\ % 
                    GCD \cite{vaze2022generalized} (CVPR-22) &  & 51.3 & 56.6 & 48.7 & 39.0 & 57.6 & 29.9 & 35.4 & 51.0 & 27.0 \\
                    \midrule
                    RIM \cite{krause2010discriminative} (NeurIPS-10) (semi-sup.)&  \cmark & 52.3 & 51.8 & 52.5 & 38.9 & 57.3 & 30.1 & 40.1 & \textbf{57.6} & 30.7 \\ 
                    TIM \cite{boudiaf2020information} (NeurIPS-20) & \cmark & 53.4 & 51.8 & 54.2 & 39.3 & 56.8 & 30.8 & 40.1 & 57.4 & 30.7 \\
                    \midrule
                    \rowcolor{Gray} PIM (proposed) & \cmark & \textbf{62.7} & \textbf{75.7} & \textbf{56.2} & \textbf{43.1} & 66.9 & \textbf{31.6} & \textbf{42.3} & 56.1 & \textbf{34.8} \\
                    \toprule
                     & & \multicolumn{3}{c}{CIFAR10} & \multicolumn{3}{c}{CIFAR100} & \multicolumn{3}{c}{ImageNet-100} \\
                     \cmidrule(r){3-5} \cmidrule(r){6-8} \cmidrule(r){9-11} Approach & InfoMax & All & Old & New & All & Old & New & All & Old & New \\
                     \midrule
                    k-means &  & 83.6 & 85.7 & 82.5 & 52.0 & 52.2 & 50.8 & 72.7 & 75.5 & 71.3 \\
                    RankStats+ \cite{han2021autonovel} (TPAMI-21) &  & 46.8 & 19.2 & 60.5 & 58.2 & 77.6 & 19.3 & 37.1 & 61.6 & 24.8 \\
                    UNO+ \cite{fini2021unified} (ICCV-21) &  & 68.6 & \textbf{98.3} & 53.8 & 69.5 & 80.6 & 47.2 & 70.3 & 95.0 & 57.9 \\
                    ORCA \cite{cao2022openworld} (ICLR-22) &  & 88.9 & 88.2 & 89.2 & 55.1 & 65.5 & 34.4 & 67.6 & 90.9 & 56.0 \\ % ResNet50
                    ORCA \cite{cao2022openworld} - ViTB16 &  & \textbf{97.1} & 96.2 & \textbf{97.6} & 69.6 & 76.4 & 56.1 & 76.5 & 92.2 & 68.9 \\ % 
                    GCD \cite{vaze2022generalized} (CVPR-22) &  & 91.5 & 97.9 & 88.2 & 70.8 & 77.6 & 57.0 & 74.1 & 89.8 & 66.3 \\
                    \midrule
                    RIM \cite{krause2010discriminative} (NeurIPS-10) (semi-sup.)&  \cmark & 92.4 & 98.1 & 89.5 & 73.8 & 78.9 & 63.4 & 74.4 & 91.2 & 66.0 \\ 
                    TIM \cite{boudiaf2020information} (NeurIPS-20) & \cmark & 93.1 & 98.0 & 90.6 & 73.4 & 78.3 & 63.4 & 76.7 & 93.1 & 68.4 \\ 
                    \midrule
                    \rowcolor{Gray} PIM (proposed) & \cmark & 94.7 & 97.4 & 93.3 & \textbf{78.3} & \textbf{84.2} & \textbf{66.5} & \textbf{83.1} & \textbf{95.3} & \textbf{77.0} \\
                    \bottomrule
                \end{tabular}
                }
                \end{sc}
                \end{small}
                \end{center}
                %\vskip -0.15in
                \caption{\textbf{Generalized Category Discovery partitioning.} Partitioning ACC scores across fine-grained and generic datasets.
                }
                \label{tab_accv2_all_dsets}
\end{table*}

\subsection{Experiments setting} \label{subsec_exp_setting}

\paragraph{Datasets.}We evaluate and compare our approach to GCD state-of-the-art approaches across six different natural image datasets. More concretely, this includes three well-known generic object recognition datasets CIFAR10 \cite{krizhevsky2009learning}, CIFAR100 \cite{krizhevsky2009learning}, ImageNet-100 \cite{deng2009imagenet}, as well as the recent semantic shift benchmark suite (SSB) \cite{vaze2021open} which is composed of the three fine-grained datasets CUB \cite{wah2011caltech}, Stanford Cars \cite{krause20133d} and Herbarium19 \cite{tan2019herbarium}. Note that the datasets on SSB bring an additional challenge to the performance of the baselines. CUB and Stanford Cars contain fine-grained categories, which are arguably harder to distinguish than generic object classes. Herbarium19 is a long-tailed dataset, which reflects a real-world use case with large class imbalance, and large intra-class and low inter-class variations.

We follow the original GCD setting \cite{vaze2022generalized} to split the original training set of each dataset into labeled and unlabeled subsets. More concretely, half of the image samples corresponding to the $K^{\text{old}}$ known classes are assigned to the labeled subset, whereas the remaining half are assigned to the unlabeled subset. The latter also contains all the image samples from the remaining classes present in the original dataset, which we consider as the novel classes. In this way, the unlabeled subset is composed of instances from $K$ different classes. 
%
%%%
Table \ref{tab_GCD_datasets_details} details for each dataset the selected number of classes, as well as the number of corresponding examples present in each subset, w.r.t. the generalized category discovery setting introduced in \cite{vaze2022generalized}.
\begin{table}[ht]
    \addtolength{\tabcolsep}{-3pt}
    %\vskip -0.08in
    \begin{center}
    \begin{small}
    \begin{sc}
    \resizebox{0.99\columnwidth}{!}{%
    \begin{tabular}{lcccccc}
        \toprule
         & CIFAR10 & CIFAR100 & ImageNet-100 & CUB & SCars & Herbarium19 \\
        \midrule
        $|\calY_L|$ & 5 & 80 & 50 & 100 & 98 & 341 \\
        $|\calY_U|$ & 10 & 100 & 100 & 200 & 196 & 683 \\
        \midrule
        $|\mathcal{D}_L|$ & 12.5k & 20k & 31.9k & 1.5k & 2.0k & 8.9k \\
        $|\mathcal{D}_U|$ & 37.5k & 30k & 95.3k & 4.5k & 6.1k & 25.4k \\
        \bottomrule
    \end{tabular}
    }
    \end{sc}
    \end{small}
    \end{center}
    %\vskip -0.2in
    \caption{Composition of the datasets used.}
    \label{tab_GCD_datasets_details}
    %\vskip -0.29in
\end{table}
%%%

\paragraph{Evaluation protocol.} To evaluate our model we follow the evaluation protocol presented in GCD \cite{vaze2022generalized}. In particular, for the partitioning task, we first employ the Hungarian algorithm to solve the optimal cluster-to-class assignment jointly on both known and novel class examples, i.e. for all the predicted clusters at once. Note that our partial semi-supervised training already enables to correctly align beforehand the clusters corresponding to the known classes with real-class labels. Then, we use this optimal label-assignment solution to estimate the overall partitioning accuracy (ACC) for all classes (\textsc{All}), for known classes (\textsc{Old}), and for novel classes (\textsc{New}). In the evaluation, we also report the estimated number of classes $\hat{K}$ in the unlabeled set and the corresponding error $Err=\frac{|\hat{K}-K|}{K}$, with $K$ the real number of classes for each dataset.

\newpage
\paragraph{Implementation details:}
\begin{itemize}
    \item \textbf{Encoder $g_{\bm{\theta}}$:} As in \cite{vaze2022generalized}, we employ the vision transformer ViT-B-16 \cite{dosovitskiy2021an} as our backbone encoder $g_{\bm{\theta}}$ (i.e. the feature extractor). It is first pre-trained on the unlabeled dataset ImageNet \cite{deng2009imagenet} with DINO \cite{caron2021emerging} self-supervision. Then, it is fine-tuned on each GCD dataset of interest with a semi-supervised contrastive loss composed  of an unsupervised noise contrastive term \cite{gutmann2010noise} and a supervised contrastive term \cite{khosla2020supervised}. It is empirically demonstrated in \cite{vaze2022generalized} that this pre-training procedure achieves robust feature representations. The resulting feature dimension is 768 per input image.
    \item \textbf{Partitioning model $f_{\bm{W}}$:} Our partitioning model follows the architecture of a standard linear classifier. We first initialize $f_{\bm{W}}$ prototypes $\bm{W}$ with the centroids produced by the semi-supervised k-means (ssKM) clustering model\footnote{Appendix provides a description of ssKM.} on the entire feature map set $\calZ$. The maximum number of clustering iterations for ssKM is set to $100$. Then, we train $f_{\bm{W}}$ with the standard Adam optimizer \cite{kingma2014adam}, with a learning rate of $0.001$ and a weight decay of $0.01$, for $1000$ epochs during the partitioning task, but only for $500$ epochs during the search of $\hat{K}$ in order to reduce the computational cost. We set the training batch size equal to the size of the dataset which is quite feasible in terms of memory and computation since our approach only requires the pre-computed feature maps.
    \item \textbf{Conditional entropy weight $\lambda$:} During the search of the number of classes, we simply set $\lambda=1$ for the unsupervised discriminative clustering step. However, during the partitioning task where the number of classes is fixed, i.e. with $K$ assumed to be known or to be equal to $\hat{K}$, we automatically select the optimal value for $\lambda$ in the interval $(0,1]$, as previously detailed in Sec. \ref{subsec_bilevel_optim}.
\end{itemize}

\subsection{Main results}
\label{ssec:mainRes}
In this section, we perform a comprehensive empirical evaluation of our method and compare it to GCD \cite{vaze2022generalized}, as well as several adapted state-of-the-art approaches. In particular, RankStats+ and UNO+ are the adapted versions from RankStats \cite{han2021autonovel} and UNO \cite{fini2021unified}, which were originally developed for the NCD task. Furthermore, results when applying simply \textit{k-means} \cite{macqueen1967classification} on the raw extracted features from DINO are also reported. Scores for k-means, RankStats+, UNO+ and GCD \cite{vaze2022generalized} are reported from \cite{vaze2022generalized}. We have also evaluated ORCA with its original ResNet  architecture \cite{he2016deep} by using authors code\footnote{\url{https://github.com/snap-stanford/orca}}, as well as with the more competitive ViT-B-16 architecture \cite{dosovitskiy2021an}, %for the sake of fairness, 
as both GCD \cite{vaze2022generalized} and our method resort to this architecture. %and our method which also use the latter. 
In addition, in order to better highlight the superior performance of the proposed approach compared to previous mutual information strategies, we have also adapted the InfoMax approaches RIM \cite{krause2010discriminative} and TIM \cite{boudiaf2020information}, previously discussed in Sec. \ref{sec_related_work}, to this novel setting. Specifically, TIM and the semi-supervised version of RIM were originally designed to deal with semi-labeled datasets, where both the labeled and unlabeled sets contain examples from the same classes.
Thus, we have expanded the number of prototypes in RIM and TIM, and their resulting prediction output vector from the $K^{\text{old}}$ to $K$. For the sake of fairness, we use the same training hyper-parameters and initialization prototypes as in our approach, as detailed in Sec. \ref{subsec_exp_setting}. Furthermore, we also applied RIM and TIM on top of the same fixed feature extractor, similarly to ssKM in GCD \cite{vaze2022generalized} and our approach.

\noindent 
We first focus on the partitioning task (\cref{tab_accv2_all_dsets}), which evaluates the label assignment performance ACC on the unlabeled samples. Following the original GCD setting \cite{vaze2022generalized}, we assume the real number of classes $K$ to be known.

\noindent \textbf{Comparisons with the GCD state-of-the-art.} Overall, we can observe that our approach PIM significantly outperforms the state-of-the-art methods UNO+, RankStat+ and GCD \cite{vaze2022generalized}, with a consistent performance improvement ranging from 3\% on CIFAR10 up to 11\% on CUB, when considering $\mathtt{ALL}$ categories. Compared to ORCA, one can notice that improvement gain brought by our method is particularly significant on the fine-grained datasets, where it ranges from 9.3\% on Stanford Cars, up to 24.7\% on CUB. In contrast, the performance differences are less remarkable across general datasets. In particular, PIM does not outperform ORCA-ViTB16 in the simpler CIFAR-10 dataset, whereas it brings 8-9\% in performance gains in CIFAR-100 and ImageNet-100, arguably more complex datasets. These results suggest that our approach is more suitable in scenarios where the total number of classes is relatively large, and potentially presenting a higher degree of class imbalance.

% Compare with TIM and RIM
\noindent \textbf{Comparison with adapted RIM \cite{krause2010discriminative} and TIM \cite{boudiaf2020information}.} The results obtained by the adapted semi-supervised RIM, TIM, and our approach PIM show the overall superiority of mutual information based methods compared to GCD \cite{vaze2022generalized}. In addition, one can observe that while RIM and TIM present very similar results. This behaviour could be explained by the fact that these two adaptations to the \textit{GCD} problem amount to use the same fixed loss function, and performance differences may be due to the classifier choice for the conditional model. Indeed, RIM uses a multiclass logistic regression, whereas the soft-classifier of TIM measures the l2 norm between prototypes (i.e. classifier weights) and L2-normalized embedded features. Last, and more importantly, we can observe that methods based on the standard mutual information yield suboptimal results compared to the proposed approach. We hypothesize that the reason for these differences is two-fold: (i) RIM and TIM compute the marginal entropy term exclusively over the unlabeled data, while we compute it over the entire distribution. In other words, PIM maximizes the mutual information over the whole dataset, which enables to better capture the entire data distribution; (ii) thanks to the proposed bi-level optimization process, the optimal lambda parameter for each dataset can be estimated automatically (as detailed in Section \ref{subsec_bilevel_optim}). We stress that our hypothesis is supported by empirical evidence in Section \ref{subsec_ablation_studies}.

\subsection{Ablation studies} \label{subsec_ablation_studies}

\begin{figure}%
\centering
\begin{minipage}{0.32\columnwidth}%
\centering
CUB
\end{minipage}%
\begin{minipage}{0.32\columnwidth}%
\centering
Stanford-Cars
\end{minipage}%
\begin{minipage}{0.32\columnwidth}%
\centering
Herbarium19
\end{minipage}%

\begin{minipage}{0.32\columnwidth}%
\centering
 
\end{minipage}%
\begin{minipage}{0.32\columnwidth}%
\centering
 
\end{minipage}%
\begin{minipage}{0.32\columnwidth}%
\centering
 
\end{minipage}%

\begin{minipage}{0.32\columnwidth}%
\includegraphics[width=\columnwidth]{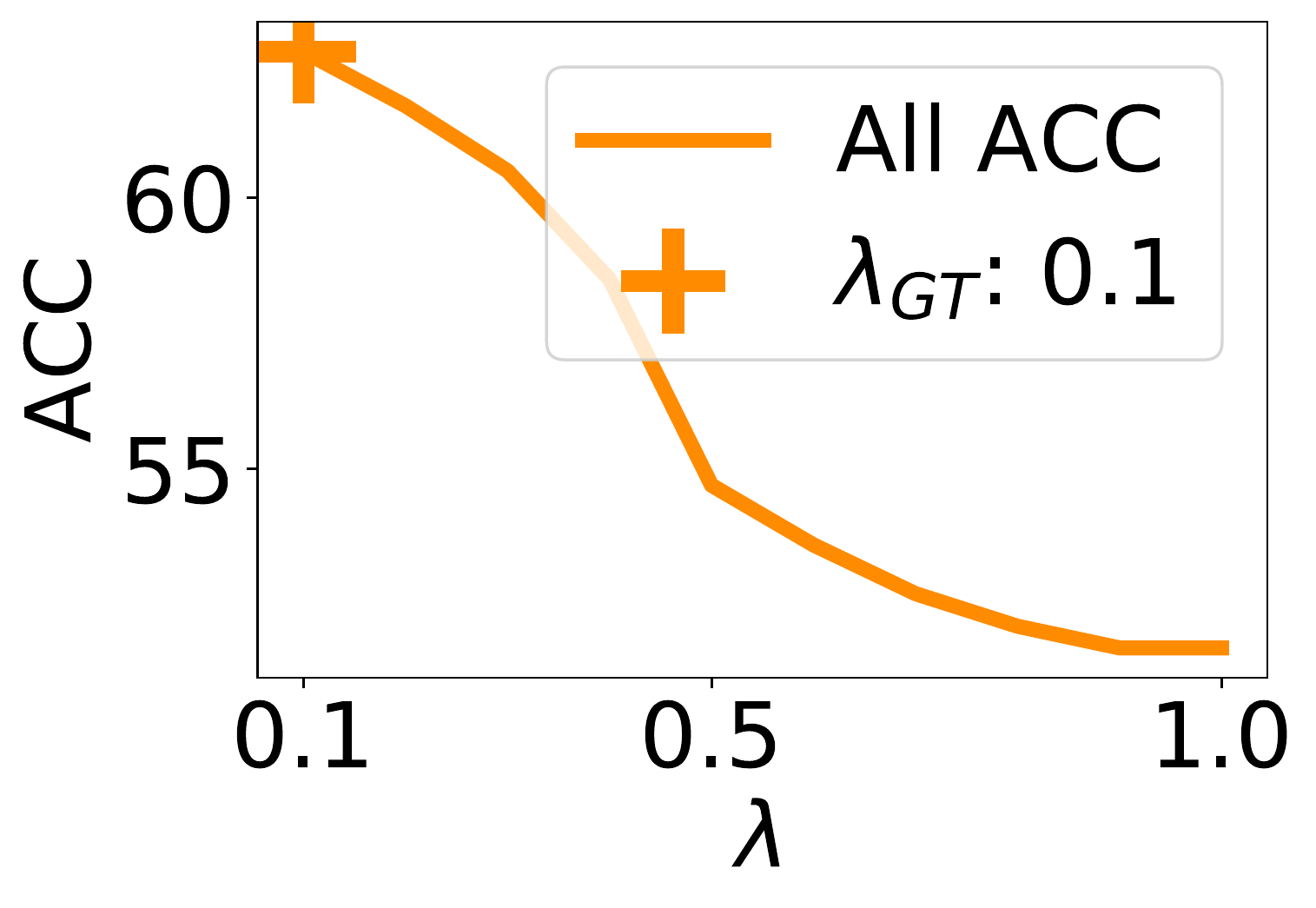}%
\centering
\\(a) 
\end{minipage}%
\begin{minipage}{0.32\columnwidth}%
\includegraphics[width=\columnwidth]{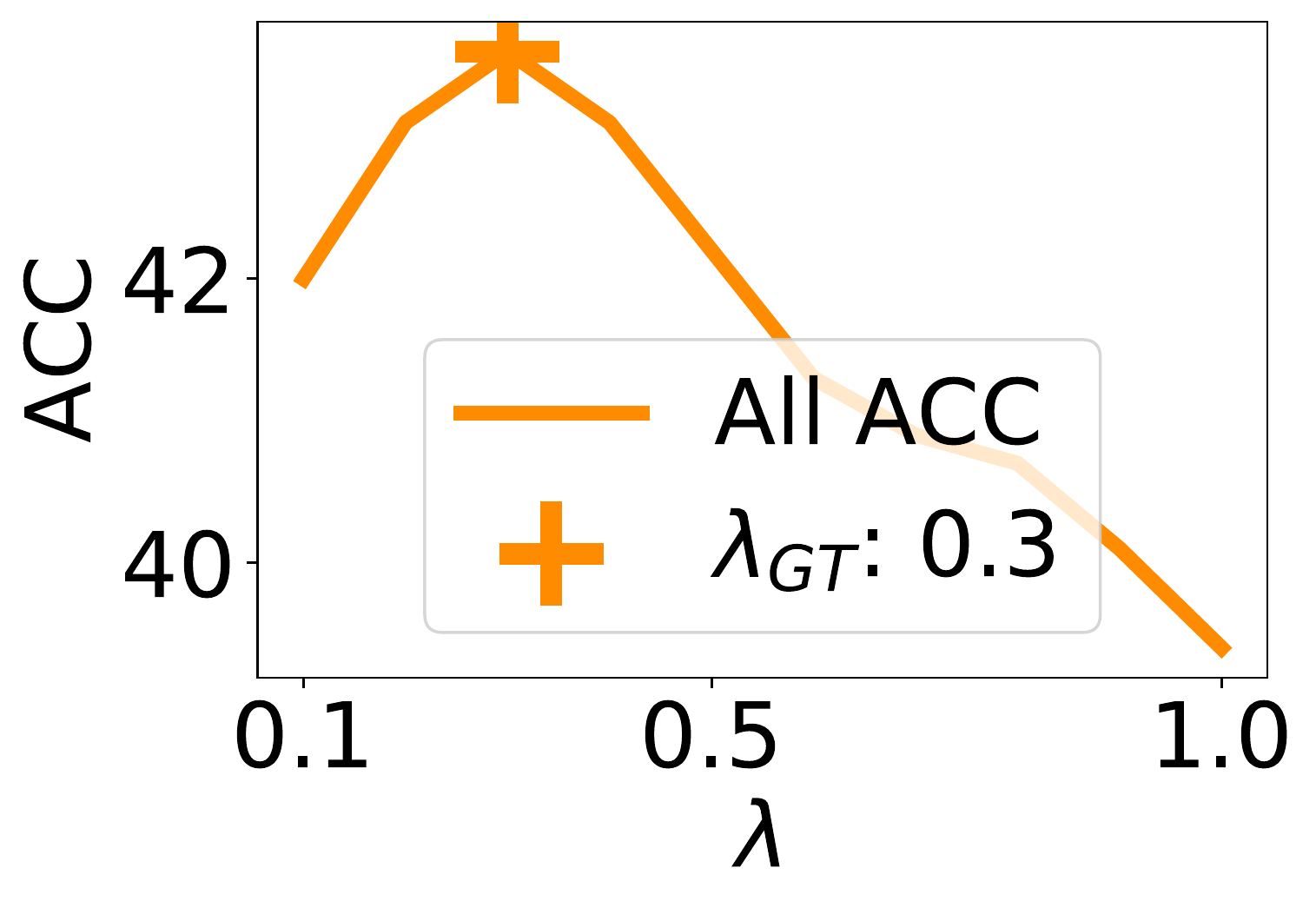}%
\centering
\\(b) 
\end{minipage}%
\begin{minipage}{0.32\columnwidth}%
\includegraphics[width=\columnwidth]{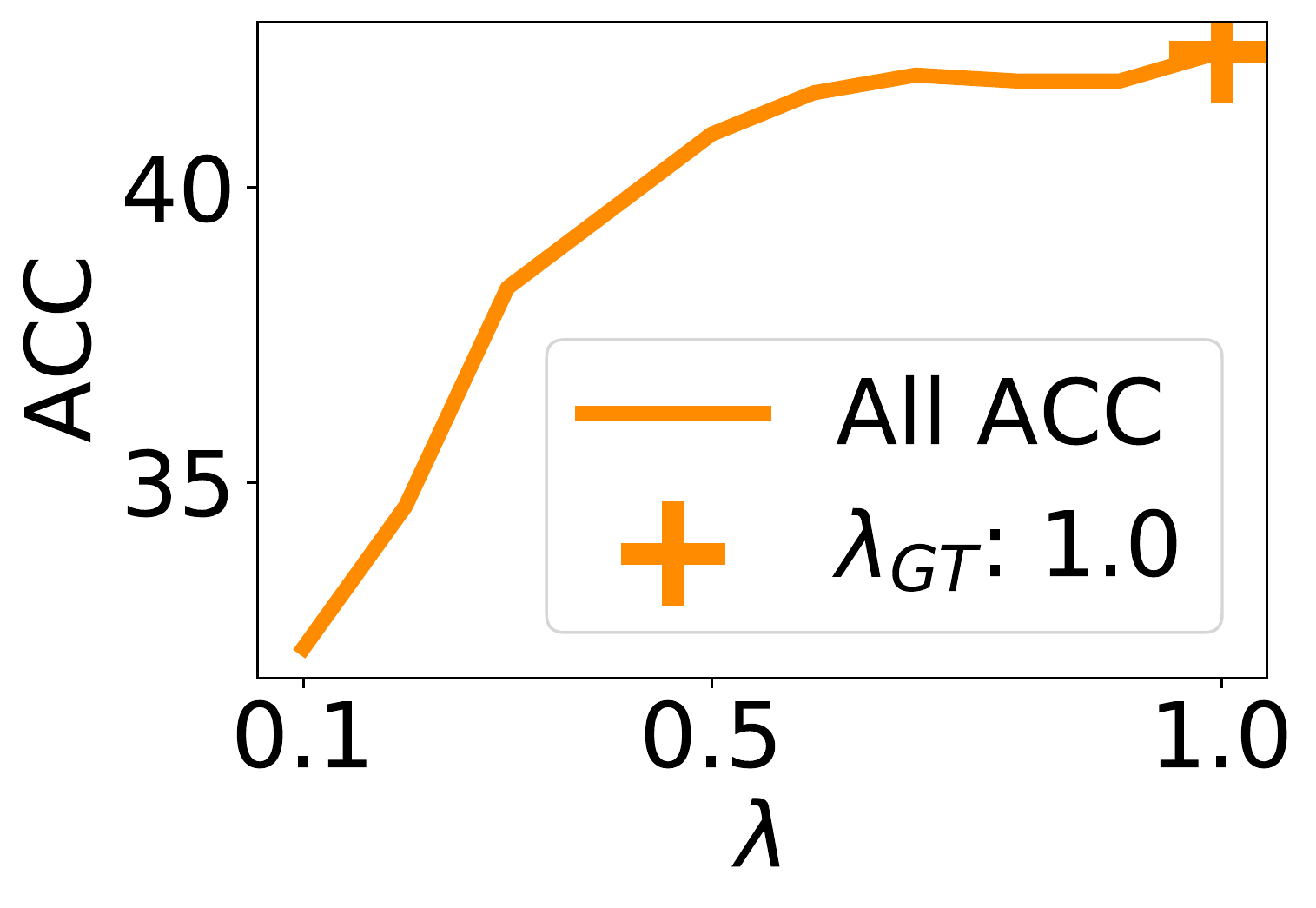}%
\centering
\\(c) 
\end{minipage}%

\begin{minipage}{0.32\columnwidth}%
\includegraphics[width=\columnwidth]{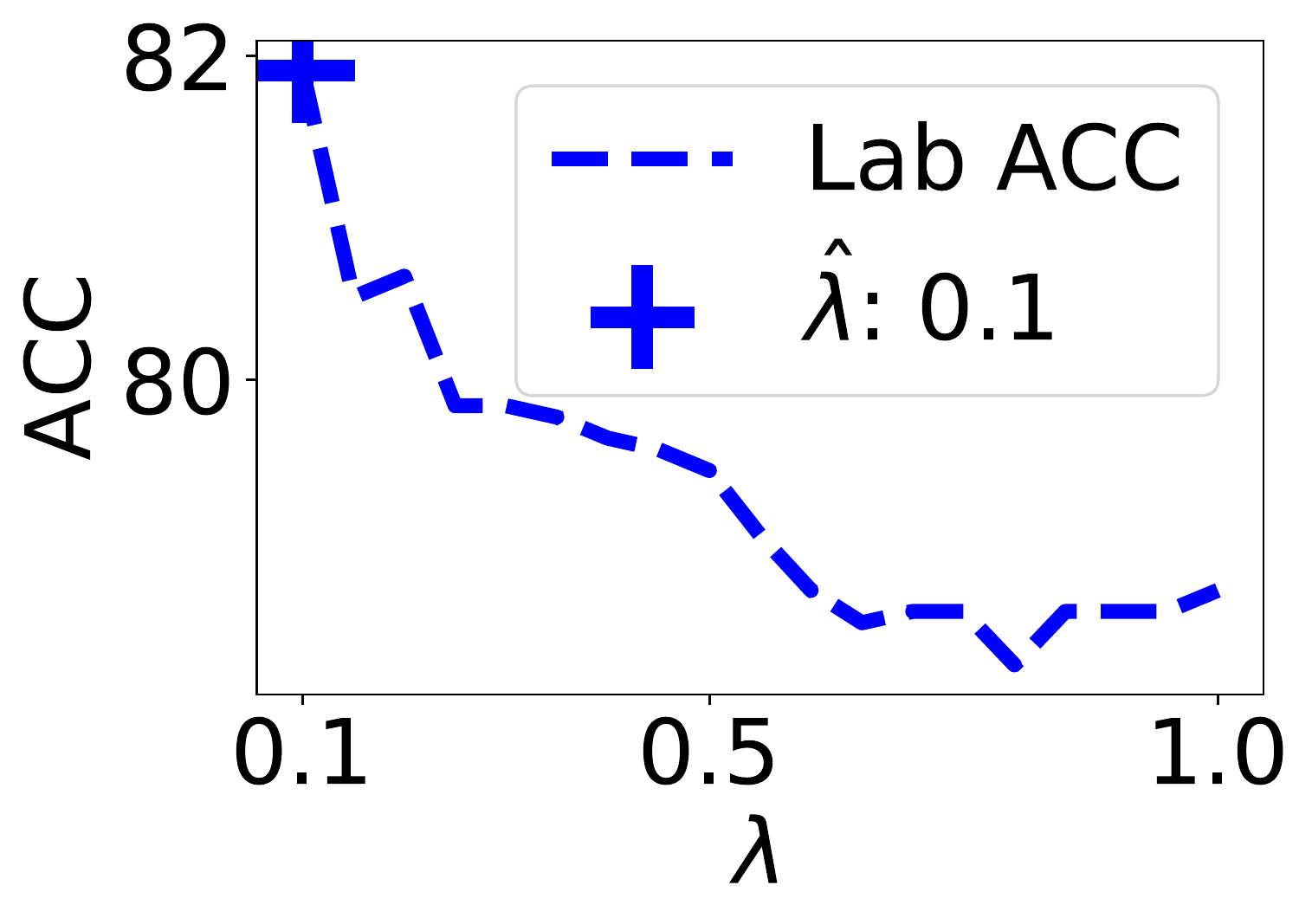}%
\centering
\\(d) 
\end{minipage}%
\begin{minipage}{0.32\columnwidth}%
\includegraphics[width=\columnwidth]{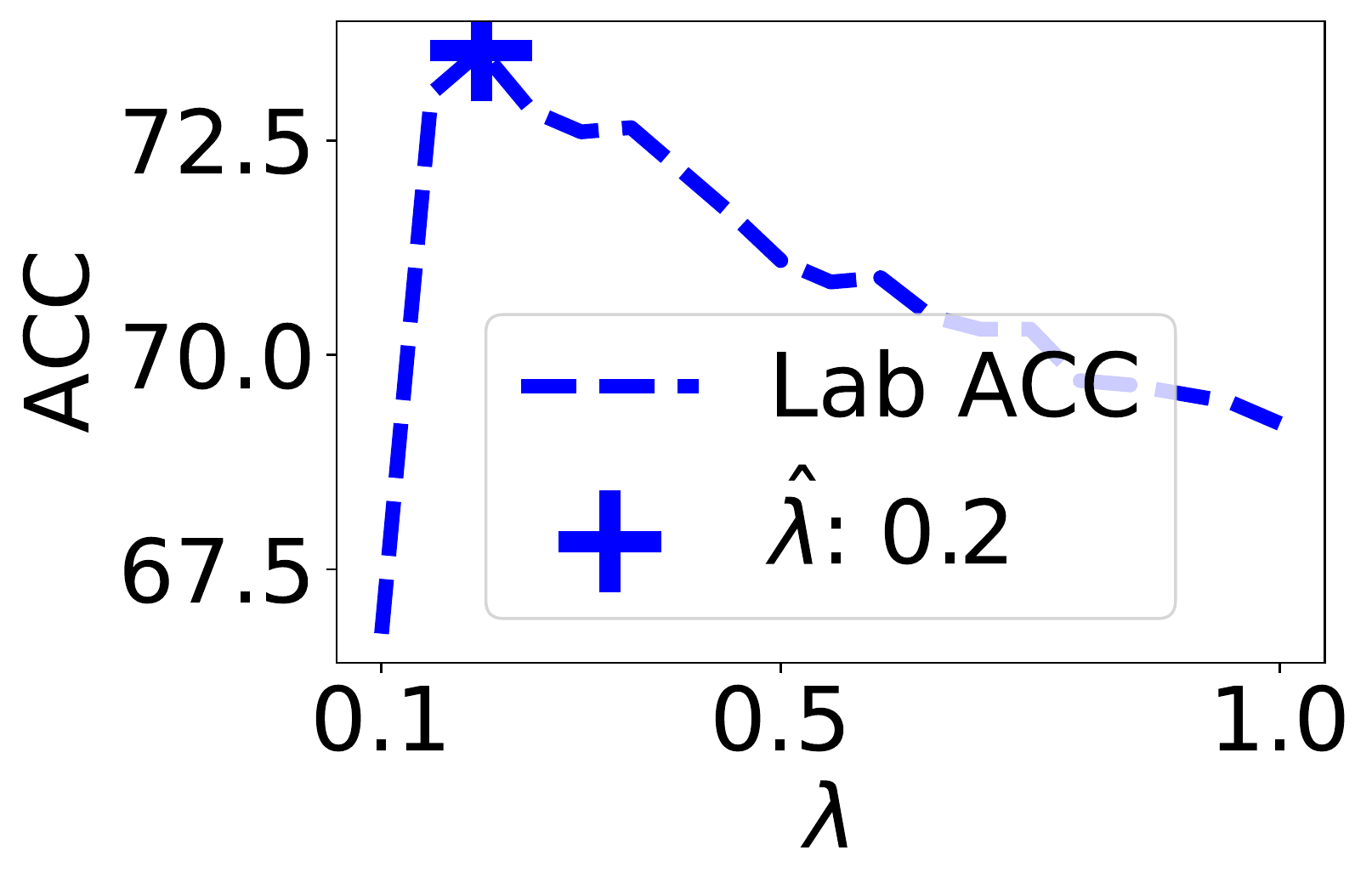}%
\centering
\\(e) 
\end{minipage}%
\vspace{0.015\textwidth}
\begin{minipage}{0.32\columnwidth}%
\includegraphics[width=\columnwidth]{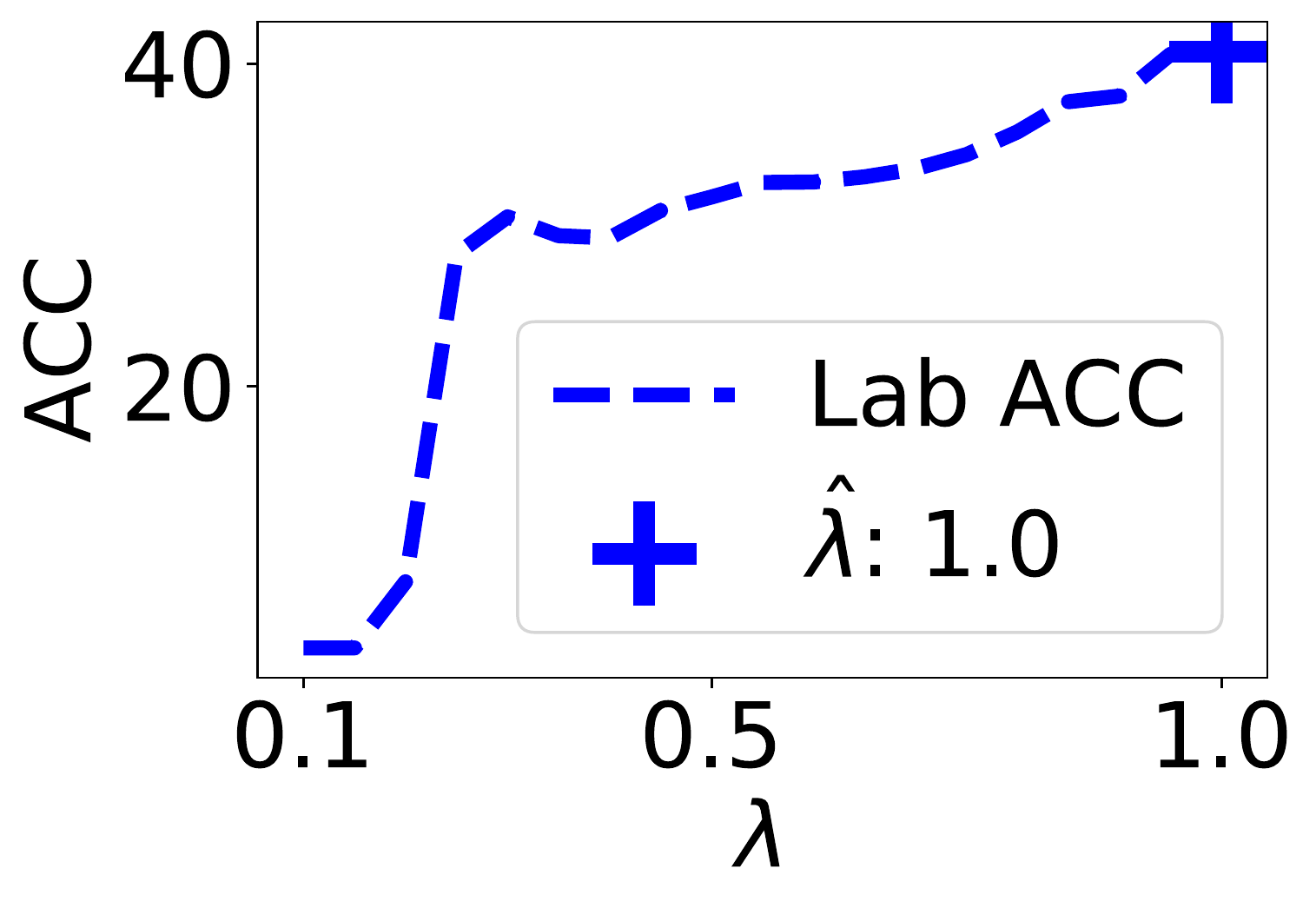}%
\centering
\\(f) 
\end{minipage}%

\begin{minipage}{0.33\columnwidth}%
\centering
 
\end{minipage}%
\begin{minipage}{0.33\columnwidth}%
\centering
 
\end{minipage}%
\begin{minipage}{0.33\columnwidth}%
\centering
 
\end{minipage}%

\begin{minipage}{0.33\columnwidth}%
\includegraphics[width=\columnwidth]{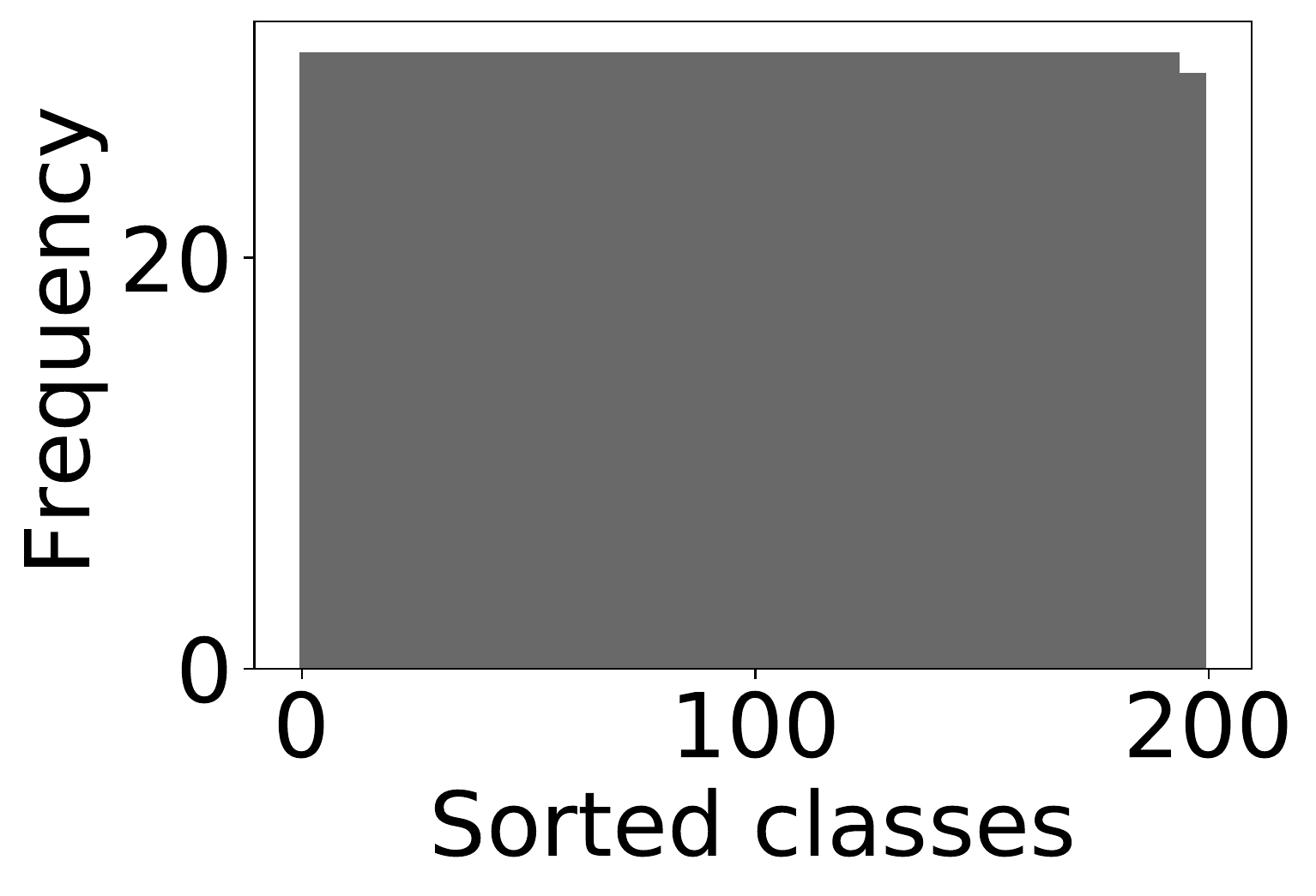}%
\centering
\\(g) 
\end{minipage}%
\begin{minipage}{0.33\columnwidth}%
\includegraphics[width=\columnwidth]{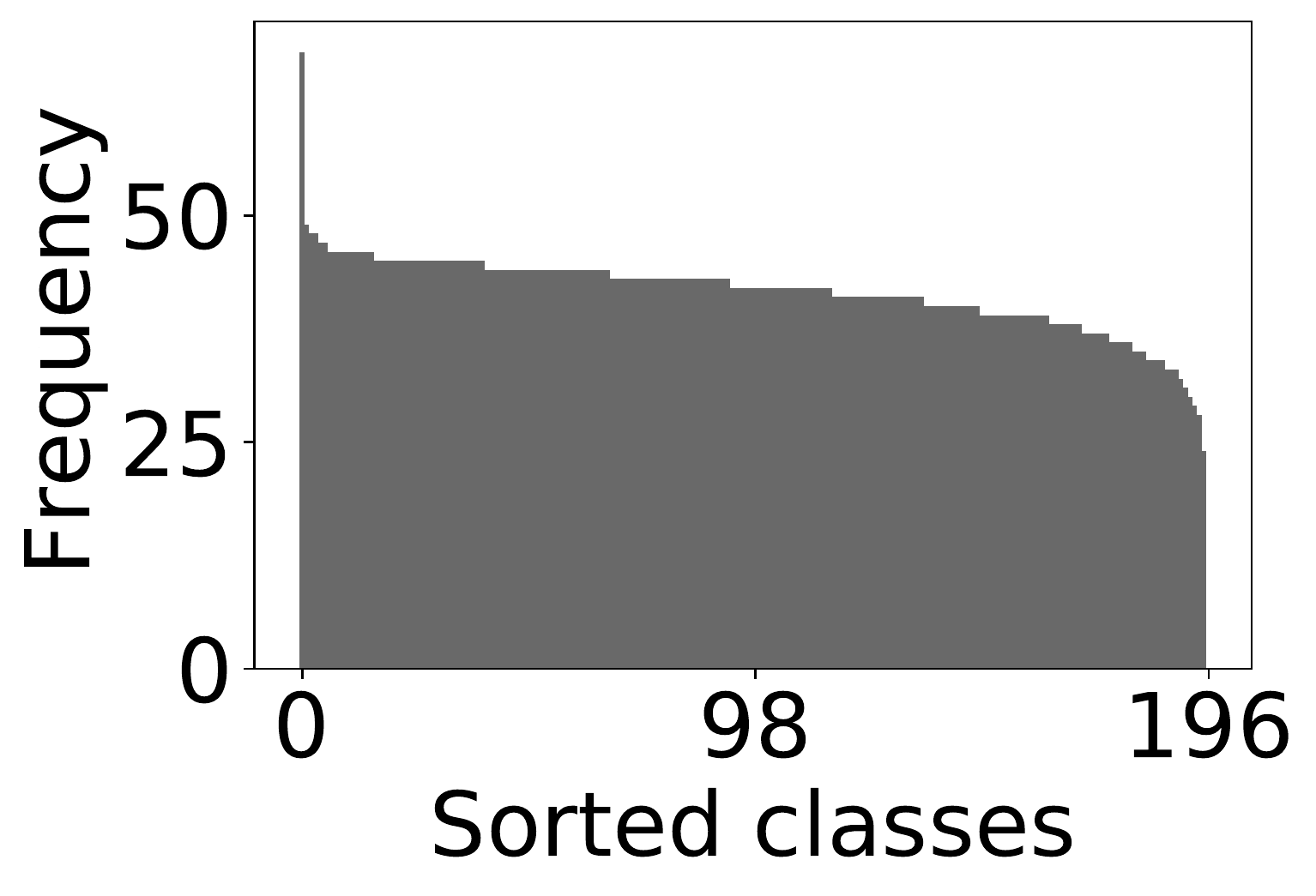}%
\centering
\\(h) 
\end{minipage}%
\begin{minipage}{0.33\columnwidth}%
\includegraphics[width=\columnwidth]{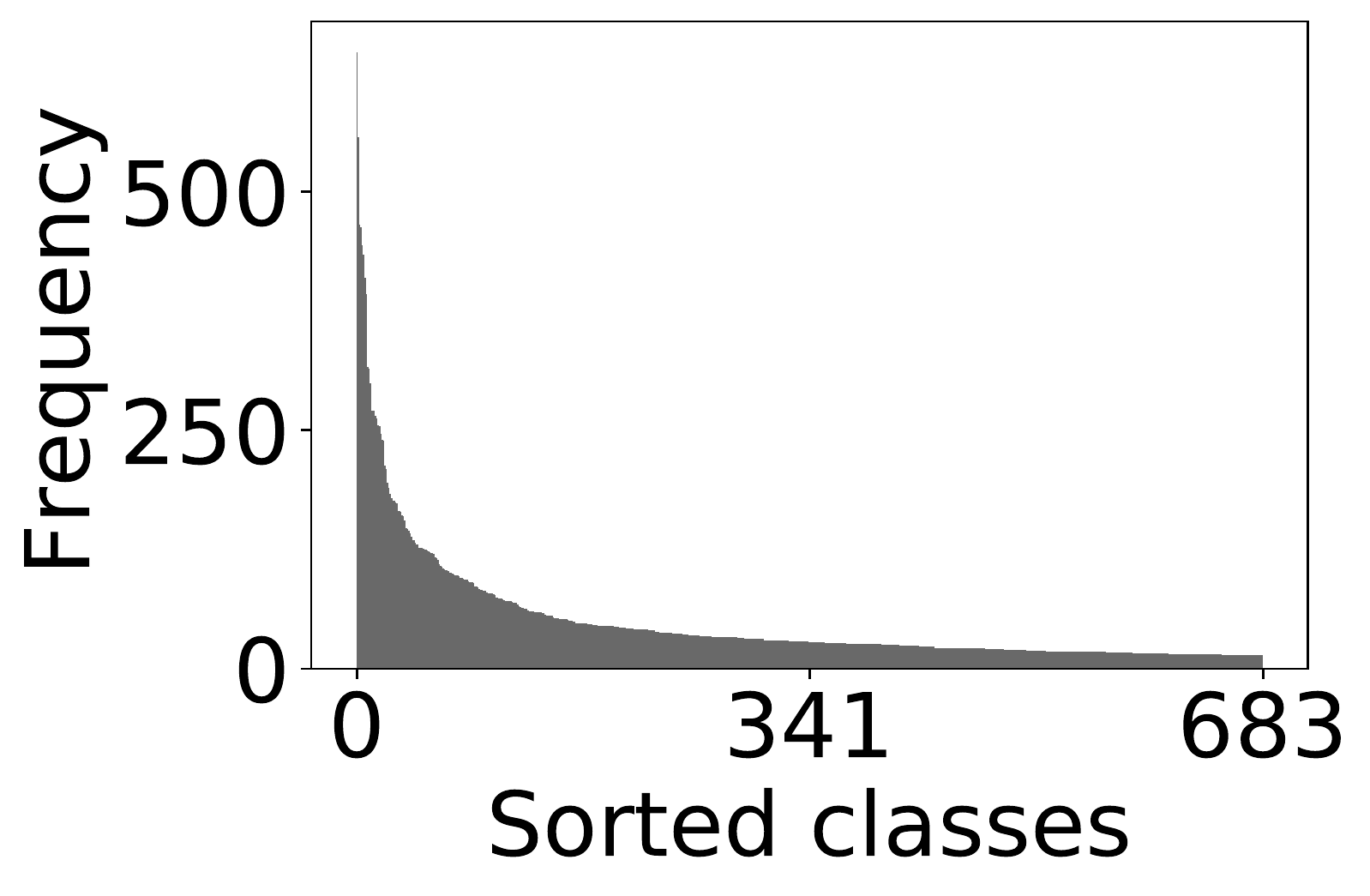}%
\centering
\\(i) 
\end{minipage}%
%\vskip -0.05in
\vskip 0.15in
\caption{\textbf{$\lambda$ effect analysis on fine-grained datasets.} The first row represents the ACC on the labeled points depending on $\lambda$ value. The second row represents the ACC on all the unlabeled points depending on $\lambda$ value. The third row represents the frequency of examples per class, in a sorted order.
}%
\label{fig_empirical_lambda_finding}%
\vskip -0.15in
\end{figure}

\begin{table*}[ht]
 \addtolength{\tabcolsep}{-3pt}
                %\vskip -0.55in
                \begin{center}
                \begin{small}
                \begin{sc}
                \resizebox{0.9\textwidth}{!}{%
                \begin{tabular}{cccccccccccc}
                    \toprule
                      & & & \multicolumn{3}{c}{CUB} & \multicolumn{3}{c}{Stanford Cars} & \multicolumn{3}{c}{Herbarium19}  \\
                    \cmidrule(r){4-6} \cmidrule(r){7-9} \cmidrule(r){10-12} Loss terms used & ${\mathcal{H}}(Y)$ on & s.t. $\bm{y}_i=\bm{p}_i$ $\forall \bm{z}_i \in \calZ_L$ & All & Old & New & All & Old & New & All & Old & New \\
                    \midrule
                    $-{\mathcal{H}}(Y | Z)$ & not used & \xmark & 6.1 & 0.0 & 9.1 & 2.5 & 0.0 & 3.6 & 4.0 & 4.0 & 4.1 \\
                    $-{\mathcal{H}}(Y|Z)$ & not used & \cmark & 38.6 & 46.2 & 34.8 & 29.8 & 51.3 & 19.5 & 34.6 & 45.4 & 28.9 \\
                    ${\mathcal{H}}(Y)-{\mathcal{H}}(Y | Z)$ & $\calZ_U$ & \xmark & 53.7 & 55.5 & 52.9 & 37.4 & 49.5 & 31.6 & 35.2 & 39.0 & 33.2 \\
                    ${\mathcal{H}}(Y)-{\mathcal{H}}(Y | Z)$ & $\calZ=\calZ_L \cup \calZ_U$ & \xmark & 56.6 & 66.4 & 51.7 & 40.8 & 60.2 & 31.5 & 36.1 & 41.0 & 33.4 \\
                    %\midrule
                    ${\mathcal{H}}(Y)-{\mathcal{H}}(Y | Z)$ & $\calZ_U$ & \cmark & 58.3 & 72.8 & 51.1 & 41.4 & 66.6 & 29.2 & 40.1 &  \textbf{57.3} &  30.8 \\
                    ${\mathcal{H}}(Y)-{\mathcal{H}}(Y | Z)$ & $\calZ=\calZ_L \cup \calZ_U$ & \cmark & \textbf{62.7} & \textbf{75.7} & \textbf{56.2} & \textbf{43.1} & \textbf{66.9} & \textbf{31.6} & \textbf{42.3} & 56.1 & \textbf{34.8} \\
                    \bottomrule
                \end{tabular}
                }
                \end{sc}
                \end{small}
                \end{center}
                \vskip -0.05in
                \caption{\textbf{Loss terms and constraint effects} on prediction performances (ACC) of proposed method PIM on fine-grained datasets.}     
                \label{tab_loss_terms_effect_fine_grained_dsets}
                %\vskip -0.1in
\end{table*}

Along these ablation studies, we focus the attention on the challenging fine-grained datasets, as the interest for each of our technical choices can be better observed.

\subsubsection{Automatic finding of optimal $\lambda$: handling both short-tailed and long-tailed datasets} We now motivate the interest of estimating the most appropriate $\lambda$ value for each dataset by using the proposed automatic finding strategy presented in \cref{subsec_bilevel_optim}. $\lambda_{GT}$ in the Figures (a), (b), (c) is the $\lambda$ value obtained when using the ground-truth labels, i.e. the value which provides the maximum performance on the unlabeled set, and $\hat{\lambda}$ in Figures (d), (e), (f) is the estimated optimal value.
Interestingly, Figures \ref{fig_empirical_lambda_finding} (a), (b), (c) first show how much the performance $\all$ ACC can significantly vary depending on the selected $\lambda$, hence motivating the interest of finding the optimal $\lambda$ value for each dataset. %
Second, Figures \ref{fig_empirical_lambda_finding} (d), (e), (f) validate our hypothesis that selecting the $\lambda$ value that maximizes the Lab ACC on the labeled data in our unconstrained conditional version, also maximizes the ACC on all the unlabeled data ($\all$ ACC) when using the constrained version instead, showing the correlation between both. 
Third, w.r.t the classes frequencies observed on each dataset (See Figures \ref{fig_empirical_lambda_finding} (g), (h), (i)), it is also interesting to note that a small $\lambda$ value provides better results on a dataset with uniform class distributions such as CUB, whereas a higher $\lambda$ value is more appropriate on the imbalanced dataset Herbarium19. Indeed, $\lambda$ controls the relative effect of the marginal entropy term in \eqref{eq_classifier_loss}. Thus, these results show that automatically selecting $\lambda$ can mitigate the encoded bias of the standard mutual information towards balanced partitions by giving more importance to the conditional entropy term in long-tailed (imbalanced) scenarios.  

\subsubsection{Effect of each loss term} We now evaluate the contribution of each term in our learning objective \eqref{eq_classifier_loss}. In particular, \cref{tab_loss_terms_effect_fine_grained_dsets} highlights the corresponding effect of the conditional entropy, the marginal entropy, and the proposed penalty constraint (i.e. replacing conditional entropy with CE on labeled points). From these results, we can draw three different observations: 1) Only minimizing the conditional entropy term produces degenerated solutions, as expected. 2) Maximizing the marginal entropy term, as well as enforcing the proposed constraint, prevents these undesired degenerated solutions. 3) Maximizing the marginal entropy term on the entire dataset, and hence maximizing as well the mutual information on the entire dataset, as we propose, further enhances the performance.

\subsection{Towards a practical setting}

\subsubsection{Estimating the number of classes} 

\begin{table}[h!]
                \addtolength{\tabcolsep}{-3pt}
                %\vskip -0.5in
                \begin{center}
                \begin{small}
                \begin{sc}
                \resizebox{0.99\columnwidth}{!}{%
                \begin{tabular}{lccccc}
                    \toprule
                    & CUB & Stanford Cars & Herbarium19 & mean \\
                     \cmidrule(r){2-2} \cmidrule(r){3-3} \cmidrule(r){4-4} \cmidrule(r){5-5} & $\hat{K}$($Err$) & $\hat{K}$($Err$) & $\hat{K}$($Err$) & ($Err$) \\ 
                    \midrule
                    Ground truth & 200(-) & 196(-) & 683(-) & (-) \\ 
                    \midrule
                    Max-ACC (GCD) \cite{vaze2022generalized} & 231 (16\%) & 230 (15\%) & 520 (24\%) & (18\%) \\ 
                    \midrule
                    \rowcolor{Gray} Max-ACC (PIM) & \textbf{227 (14\%)} & \textbf{169 (13\%)} & \textbf{563 (18\%)} & \textbf{(15\%)} \\         
                    \bottomrule
                    \toprule
                    & CIFAR10 & CIFAR100 & ImageNet-100 & mean \\
                     \cmidrule(r){2-2} \cmidrule(r){3-3} \cmidrule(r){4-4} \cmidrule(r){5-5} & $\hat{K}$($Err$) & $\hat{K}$($Err$) & $\hat{K}$($Err$) & ($Err$) \\ 
                    \midrule
                    Ground truth & 10 (-) & 100 (-) & 100 (-) & (-) \\ 
                    \midrule 
                    Max-ACC (GCD) \cite{vaze2022generalized} & 9 (10\%) & \textbf{100 (0\%)} & 109 (9\%) & (6\%)\\ 
                    \midrule
                    \rowcolor{Gray} Max-ACC (PIM) & \textbf{10 (0\%)} & 95 (5\%) & \textbf{102 (2\%)} & \textbf{(2\%)} \\ 
                    \bottomrule
                \end{tabular}
                }
                \end{sc}
                \end{small}
                \end{center}
                \vskip -0.05in
                \caption{\textbf{Estimation of the number of classes} in the unlabeled set using Brent's algorithm as in \cite{vaze2022generalized}. Max-ACC (GCD) \cite{vaze2022generalized} results are reported from \cite{vaze2022generalized}.}
                \vskip -0.15in
                \label{tab_estimate_hat_K}
\end{table}

In order to find the number of classes, we follow the strategy proposed in GCD \cite{vaze2022generalized} (which we refer to as Max-ACC (GCD)), but we replace the k-means clustering stage with the unconstrained (i.e. unsupervised) version of our %competitive 
method PIM, referred to as Max-ACC (PIM). %our alternative. 
The results from these methods are reported in \Cref{tab_estimate_hat_K}. Overall, the proposed combination Max-ACC (PIM) is more appropriate than Max-ACC (GCD) \cite{vaze2022generalized} on both generic and fine-grained datasets, except on CIFAR-100 where Max-ACC (GCD) \cite{vaze2022generalized} finds the real number of classes.

\subsubsection{Performance when the number of classes is unknown} While we followed the standard practices %in existing literature 
for the partitioning task in the experiments of Section \ref{ssec:mainRes}, we argue that having access to the number of expected classes is an unrealistic assumption. Thus, we now relax this assumption by repeating the partitioning experiments with the estimated value of $\hat{K}$ (See \cref{tab_estimate_hat_K}) instead of the real value $K$. For the GCD method \cite{vaze2022generalized}, we used the authors code\footnote{\url{https://github.com/sgvaze/generalized-category-discovery}}, except on CIFAR-100 where we directly reported scores from Tab. \ref{tab_accv2_all_dsets} because $\hat{K}=K$. These results, which are reported on \cref{tab_accv2_using_hat_k}, demonstrate the superiority of our method even in this more challenging scenario. This suggests that our formulation serves as a more robust solution in the absence of prior knowledge about $K$ for the GCD task.

\begin{table}[ht]
 \addtolength{\tabcolsep}{-3pt}
                %\vskip -0.55in
                \begin{center}
                \begin{small}
                \begin{sc}
                \resizebox{0.99\columnwidth}{!}{%
                \begin{tabular}{lccccccccc}
                    \toprule
                     & \multicolumn{3}{c}{CUB} & \multicolumn{3}{c}{Stanford Cars} & \multicolumn{3}{c}{Herbarium19} \\
                    \cmidrule(r){2-4} \cmidrule(r){5-7} \cmidrule(r){8-10} & All & Old & New & All & Old & New & All & Old & New \\
                    \midrule
                    GCD \cite{vaze2022generalized} & 51.1 & 56.4 & 48.4 & 39.1 & 58.6 & 29.7 & 37.2 & 51.7 & 29.4 \\
                    \midrule
                    \rowcolor{Gray} PIM & \textbf{62.0} & \textbf{75.7} & \textbf{55.1} & \textbf{42.4} & \textbf{65.3} & \textbf{31.3} & \textbf{42.0} & \textbf{55.5} & \textbf{34.7} \\
                    \bottomrule
                    \toprule
                     & \multicolumn{3}{c}{CIFAR10} & \multicolumn{3}{c}{CIFAR100} & \multicolumn{3}{c}{ImageNet-100}  \\
                    \cmidrule(r){2-4} \cmidrule(r){5-7} \cmidrule(r){8-10} & All & Old & New & All & Old & New & All & Old & New \\
                    \midrule
                    GCD \cite{vaze2022generalized} & 80.5 & \textbf{97.9} & 71.8 & 70.8 & 77.6 & 57.0 & 77.9 & 91.1 & 71.3 \\
                    \midrule
                    \rowcolor{Gray} PIM & \textbf{94.7} & 97.4 & \textbf{93.3} & \textbf{75.6} &  \textbf{81.6} &  \textbf{63.6} & \textbf{83.0} & \textbf{95.3} & \textbf{76.9} \\
                    \bottomrule
                \end{tabular}
                }
                \end{sc}
                \end{small}
                \end{center}
                \vskip -0.05in
                \caption{\textbf{Realistic GCD partitioning.} ACC scores are obtained by assuming $\hat{K}$ (See Tab. \ref{tab_estimate_hat_K}) as the number of expected classes.}
                \label{tab_accv2_using_hat_k}
                \vskip -0.15in
\end{table}

\section{Conclusion}

In this work, we propose a simple yet effective alternative for Generalized Category Discovery. In particular, we introduce a parametric family of mutual information objectives, which we tackle with a bi-level optimization formulation. Our solution allows to estimate the relative weight of the marginal-entropy term automatically, which mitigates the class-balance bias inherent in standard information maximization. Our empirical validation demonstrates that by learning the optimal weight that controls the relative effect of the marginal-entropy, our model deals effectively with both short-tailed and long-tailed datasets. Indeed, our presented formulation achieves new state-of-the-art results in GCD tasks, outperforming existing solutions across the different %short-tailed and long-tailed 
benchmarks by a significant margin. Moreover, in contrast to prior work, we relax the assumption that the number of classes in the unlabeled set is given, which facilitates the robustness of the proposed model in this more challenging scenario, and its scalability to more realistic settings. Furthermore, it is noteworthy to mention that our formulation is flexible and could, therefore, be coupled with any trained feature extractor. Thus, we hope that the proposed framework will be useful for future research and development to solve the GCD problem for real-world applications.

\textbf{Limitations.} A common limitation in the current GCD learning paradigm stems from the fact that the models require access to the entire target unlabeled dataset at test time. Needless to say, this strong assumption might hinder the scalability of these approaches when the target set is composed of a small number of images, or when samples appear sequentially. As in \cite{vaze2022generalized}, we assume that optimal values for finding $\lambda$ and $K$ are obtained when the ACC is maximized on the available labeled points. Despite that our extensive experiments empirically confirm that this assumption is promising in practice, it could be interesting to find a metric that could simultaneously consider the novel classes. 

{\small
\bibliographystyle{ieee_fullname}
\bibliography{egbib}
}

%%%%%%%%%%%%%%%%%%%%%%%%%%%%
%%%%%%%% Appendix %%%%%%%%%%
%%%%%%%%%%%%%%%%%%%%%%%%%%%%
\newpage
\input{appendix}
%%%%%%%%%%%%%%%%%%%%%%%%%%%%
%%%%%%%% End Appendix %%%%%%
%%%%%%%%%%%%%%%%%%%%%%%%%%%%

\end{document}

%% file: appendix.tex
\appendix
\section{Appendix}

\subsection{Semi-supervised k-means (ssKM)} \label{ssKM_details}

\textbf{Parameters initialization for ssKM.} We denote the centroid parameters of ssKM as $\bm{W} = (\bm{W}^{old}, \bm{W}^{new})$,
where
$$
    \left\{
            \begin{array}{ll}
                \bm{W}^{old}=(\bm{w}_k^{old})_{1 \leq k \leq K^{old}} \\
                \bm{W}^{new}= (\bm{w}_k^{new})_{K^{old}+1 \leq k \leq K^{old}+K^{new}}.
            \end{array}
    \right.
$$
Here, $\bm{W}^{old}$ and $\bm{W}^{new}$ are the centroids for the known and novel classes, respectively.
We initialize $\bm{W}$ by using both the labeled set and the unlabeled set jointly. As in \cite{vaze2022generalized}, we first produce the centroids for the known classes using the labeled set, such that:
$
    \bm{w}_k^{old} = (\sum_{\bm{z}_i \in \mathcal{Z}_L} y_{i,k} \bm{z}_i) / \sum_{\bm{z}_i \in \mathcal{Z}_L} y_{i,k}.
$
Then, we initialize the centroids corresponding to the novel classes with kmeans++ initialization.% \cite{Arthur07}.

\textbf{ssKM objective and clustering process.} As in \cite{vaze2022generalized}, we update the cluster centroids of ssKM algorithm, with the following objective:
\begin{equation}
    \begin{aligned}
        L_{\text{ssKM}}({\mathbf Y};{\mathbf U};\bm{W}) &= \left( \sum_{k=1}^K \sum_{\bm{z}_i \in \mathcal{Z}_L}  y_{i,k} \|\bm{z}_i- \bm{w}_{k}\|_2 \right) \\
        &+ \left(\sum_{k=1}^K \sum_{\bm{z}_i \in \mathcal{Z}_U}  u_{i,k} \|\bm{z}_i- \bm{w}_{k}\|_2 \right), 
    \end{aligned}
    \label{eq_ssKM}
\end{equation}
where $\|.\|_2$ denotes the Euclidean distance, and $\bm{u}_i = (u_{i,k})_{1 \leq k \leq K}$ denotes the latent binary vector  assigning point $\bm{z}_i$ to cluster $k$. $\mathbf U \in \{0, 1\}^{NK}$ denotes the latent assignment matrix composed of the latent binary vectors. ssKM algorithm proceeds with the same block-coordinate descent approach as the standard unsupervised k-means \cite{macqueen1967classification}. Thus, in order to minimize $L_{\text{ssKM}}$, it proceeds with the following cluster assignment and centroid updates cycle:
\begin{itemize}
    \item {\bf U}-update: Do the label assignment of the unlabeled points such that,  
        $$u_{i,k} = \left\{
            \begin{array}{ll}
                1 & \mbox{if } \argminC_{k} \|\bm{z}_i- \bm{w}_{k}\|_2 = k \\
                0 & \mbox{otherwise.}
            \end{array}
        \right.$$
    \item $\bm{W}$-update: Find $\argmin_{\bm{W}}  L_{\text{ssKM}}({\mathbf Y};{\mathbf U};\bm{W})$
\end{itemize}
%%%
Note that the latent label assignment update step is not applied on the labeled points, for which we simply keep using the available ground-truth labels in the first term in \eqref{eq_ssKM}, all along the clustering process.

\subsection{Supplementary results}

\textbf{Effect of prototypes initialization.} 
All along our article experiments, we found empirically consistent to use ssKM (detailed in Sec. \ref{ssKM_details}) centroids to initialize the prototypes of the proposed partitioning model PIM. In this section, in order to endorse this choice, we empirically emphasize across Tab. \ref{tab_ctrds_init_fine_grained_dsets} the effect of prototypes initialization on PIM prediction performances. We thus compare the following three different possible initializations (\textsc{init}):
-\textsc{ssRDM} \textsc{init} consists of using labeled points for prototypes of known classes, and random points for prototypes of novel classes; -\textsc{ssKM++} \textsc{init} consists of using labeled points for prototypes of known classes, and kmeans++ points for prototypes of novel classes (w.r.t. known class prototypes); -\textsc{ssKM} \textsc{init} consists of using ssKM centroids as \textsc{init} prototypes. The results observed on Tab. \ref{tab_ctrds_init_fine_grained_dsets} show that the proposed approach is overall almost insensitive to prototypes initialization. One can also note that PIM performances are overall slightly improved when using directly \textsc{ssKM++} \textsc{init}.
\begin{table}[ht]
 \addtolength{\tabcolsep}{-3pt}
                %\vskip 0.15in
                \vskip -0.05in
                \begin{center}
                \begin{small}
                \begin{sc}
                \resizebox{0.99\columnwidth}{!}{%
                \begin{tabular}{lccccccccc}
                    \toprule
                      & \multicolumn{3}{c}{CUB} & \multicolumn{3}{c}{Stanford Cars} & \multicolumn{3}{c}{Herbarium19}  \\
                    \cmidrule(r){2-4} \cmidrule(r){5-7} \cmidrule(r){8-10} Init & All & Old & New & All & Old & New & All & Old & New \\
                    ssrdm & 62.1 & 76.2 & 55.1 & 43.0 & 64.3 & 32.7 & 42.7 & 55.3 & 36.0 \\
                    ssKM++ & 64.3 & 76.3 & 58.4 & 43.5 & 65.6 & 32.8 & 42.3 & 55.3 & 35.4 \\
                    %\midrule
                    ssKM & 62.7 & 75.7 & 56.2 & 43.1 & 66.9 & 31.6 & 42.3 & 56.1 & 34.8 \\
                    \bottomrule
                    
                    \toprule
                      & \multicolumn{3}{c}{CIFAR10} & \multicolumn{3}{c}{CIFAR100} & \multicolumn{3}{c}{ImageNet-100}  \\
                    \cmidrule(r){2-4} \cmidrule(r){5-7} \cmidrule(r){8-10} init & All & Old & New & All & Old & New & All & Old & New \\
                    ssrdm & 94.6 & 97.4 & 93.2 & 78.2 & 85.4 & 64.0 & 82.0 & 95.4 & 75.2 \\
                    ssKM++ & 94.7 & 97.4 & 93.3 & 78.5 & 84.3 & 66.9 & 83.6 & 95.4 & 77.8 \\
                    %\midrule
                    ssKM  & 94.7 & 97.4 & 93.3 & 78.3 & 84.2 & 66.5 & 83.1 & 95.3 & 77.0 \\
                    \bottomrule
                \end{tabular}
                }
                \end{sc}
                \end{small}
                \end{center}
                \vskip -0.1in
                \caption{\textbf{Prototypes initialization effect} on PIM ACC performances on fine-grained and generic datasets.}
                \label{tab_ctrds_init_fine_grained_dsets}
                \vskip -0.05in
\end{table}

\textbf{Effect of PIM objective components on generic datasets.} Tab. \ref{tab_loss_terms_effect_generic_dsets} shows the effect of PIM objective components on the generic datasets, in particular on \textsc{CIFAR100} and \textsc{ImageNet-100}.
\begin{table}[ht]
 \addtolength{\tabcolsep}{-3pt}
                %\vskip 0.15in
                \vskip -0.05in
                \begin{center}
                \begin{small}
                \begin{sc}
                \resizebox{0.99\columnwidth}{!}{%
                \begin{tabular}{cccccccccccc}
                    \toprule
                       &  & & \multicolumn{3}{c}{CIFAR10} & \multicolumn{3}{c}{CIFAR100} & \multicolumn{3}{c}{ImageNet-100}  \\
                    \cmidrule(r){4-6} \cmidrule(r){7-9} \cmidrule(r){10-12} Loss terms used &${\mathcal{H}}(Y)$ on & s.t. $\underbrace{\bm{y}_i=\bm{p}_i}_{\forall \bm{z}_i \in \calZ_L}$& All & Old & New & All & Old & New & All & Old & New \\
                    $-\calH(Y|Z)$ & - & $\times$ & 94.5 & 97.5 & 93.1 & 2.3 & 1.0 & 5.0 & 39.1 & 82.3 & 17.3 \\
                    $-\calH(Y|Z)$ & - & \checkmark & 94.5 & 97.5 & 93.0 & 66.3 & 73.1 & 52.8 & 54.7 & 85.6 & 39.1 \\   
                    $\calH(Y)-\calH(Y|Z)$ & $\calZ_U$ & $\times$ & 92.3 & 97.9 & 89.5 & 72.4 & 82.5 & 52.3 & 79.2 & 87.6 & 75.0 \\
                    $\calH(Y)-\calH(Y|Z)$ & $\calZ$ & $\times$ & 94.8 & 97.2 & 93.6 & 78.8 & 82.4 & 71.5 & 83.1 & 93.9 & 77.6 \\
                    %\midrule
                    $\calH(Y)-\calH(Y|Z)$ & $\calZ_U$ & \checkmark & 92.4 & 98.1 & 89.5 & 73.9 & 84.7 & 52.4 & 79.3 & 94.6 & 71.7 \\
                    $\calH(Y)-\calH(Y|Z)$ & $\calZ$ & \checkmark & 94.7 & 97.4 & 93.3 & 78.3 & 84.2 & 66.5 & 83.1 & 95.3 & 77.0 \\
                    \bottomrule
                \end{tabular}
                }
                \end{sc}
                \end{small}
                \end{center}
                \vskip -0.1in
                \caption{\textbf{PIM objective components effects} on generic datasets in terms of ACC scores.}
                \label{tab_loss_terms_effect_generic_dsets}
                \vskip -0.05in
\end{table}

\textbf{Standard deviation ($\pm$)} of PIM for the generalized category partitioning is shown on Tab.\ref{tab_std_all_dsets}.
 
\begin{table}[ht]
 \addtolength{\tabcolsep}{-3pt}
                %\vskip -0.15in
                \vskip -0.05in
                \begin{center}
                \begin{small}
                \begin{sc}
                \resizebox{0.99\columnwidth}{!}{%
                \begin{tabular}{lccccccccc}
                    \toprule
                     & \multicolumn{3}{c}{CUB} & \multicolumn{3}{c}{Stanford Cars} & \multicolumn{3}{c}{Herbarium19} \\
                    \cmidrule(r){2-4} \cmidrule(r){5-7} \cmidrule(r){8-10} Approach & All & Old & New & All & Old & New & All & Old & New \\
                    \midrule
                    PIM std.-dev. ($\pm$) & $\pm$0.5 & $\pm$0.8 & $\pm$0.7 & $\pm$0.3 & $\pm$0.9 & $\pm$0.3 & $\pm$0.3 & $\pm$0.8 & $\pm$0.2 \\
                    \bottomrule
                    
                    \toprule
                     & \multicolumn{3}{c}{CIFAR10} & \multicolumn{3}{c}{CIFAR100} & \multicolumn{3}{c}{ImageNet-100} \\
                     \cmidrule(r){2-4} \cmidrule(r){5-7} \cmidrule(r){8-10} Approach & All & Old & New & All & Old & New & All & Old & New \\
                     \midrule
                     PIM std.-dev. ($\pm$) & $\pm$0.0 & $\pm$0.0 & $\pm$0.0 & $\pm$1.0 & $\pm$0.5 & $\pm$2.8 & $\pm$0.3 & $\pm$0.0 & $\pm$0.5 \\
                    \bottomrule
                \end{tabular}
                }
                \end{sc}
                \end{small}
                \end{center}
                \vskip -0.1in
                \caption{\textbf{Standard deviation ($\pm$)} for ACC scores of PIM (Tab. \ref{tab_accv2_all_dsets}) over 5 trainings, across fine-grained and generic datasets.
                }
                \label{tab_std_all_dsets}
                \vskip -0.05in
\end{table}